\def\year{2020}\relax
\documentclass[letterpaper]{article} 
\usepackage{aaai20}  
\usepackage{times}  
\usepackage{helvet} 
\usepackage{courier}  
\usepackage[hyphens]{url}  
\usepackage{graphicx} 
\def\UrlFont{\rm}  
\usepackage{graphicx}  
\usepackage{amssymb}
\usepackage{booktabs}
\usepackage{multirow}
\usepackage{tabularx}
\newcolumntype{Y}{>{\centering\arraybackslash}X}

\title{Toward 3D Object Reconstruction from Stereo Images}
\author{%
Haozhe Xie\textsuperscript{\rm 1,2}
Hongxun Yao\textsuperscript{\rm 1}
Shangchen Zhou\textsuperscript{\rm 2}
Shengping Zhang\textsuperscript{\rm 1,4}
Xiaoshuai Sun\textsuperscript{\rm 3}
Wenxiu Sun\textsuperscript{\rm 2} \\~\\
\textsuperscript{\rm 1}Harbin Institute of Technology \hspace{2 mm}
\textsuperscript{\rm 2}SenseTime Research \hspace{2 mm}
\textsuperscript{\rm 3}Xiamen University \hspace{2 mm}
\textsuperscript{\rm 4}Peng Cheng Laboratory \\
\{hzxie,h.yao,s.zhang\}@hit.edu.cn \hspace{2 mm}
\{zhoushangchen,sunwenxiu\}@sensetime.com \hspace{2 mm}
xssun@xmu.edu.cn
}
\begin{document}

\maketitle

\begin{abstract}
Inferring the 3D shape of an object from an RGB image has shown impressive results, however, existing methods rely primarily on recognizing the most similar 3D model from the training set to solve the problem.
These methods suffer from poor generalization and may lead to low-quality reconstructions for unseen objects.
Nowadays, stereo cameras are pervasive in emerging devices such as dual-lens smartphones and robots, which enables the use of the two-view nature of stereo images to explore the 3D structure and thus improve the reconstruction performance.
In this paper, we propose a new deep learning framework for reconstructing the 3D shape of an object from a pair of stereo images, which reasons about the 3D structure of the object by taking bidirectional disparities and feature correspondences between the two views into account.
Besides, we present a large-scale synthetic benchmarking dataset, namely {\it StereoShapeNet}, containing 1,052,976 pairs of stereo images rendered from ShapeNet along with the corresponding bidirectional depth and disparity maps.
Experimental results on the {\it StereoShapeNet} benchmark demonstrate that the proposed framework outperforms the state-of-the-art methods.
\end{abstract}

\begin{figure}[!t]
  \resizebox{\linewidth}{!} {
    \includegraphics{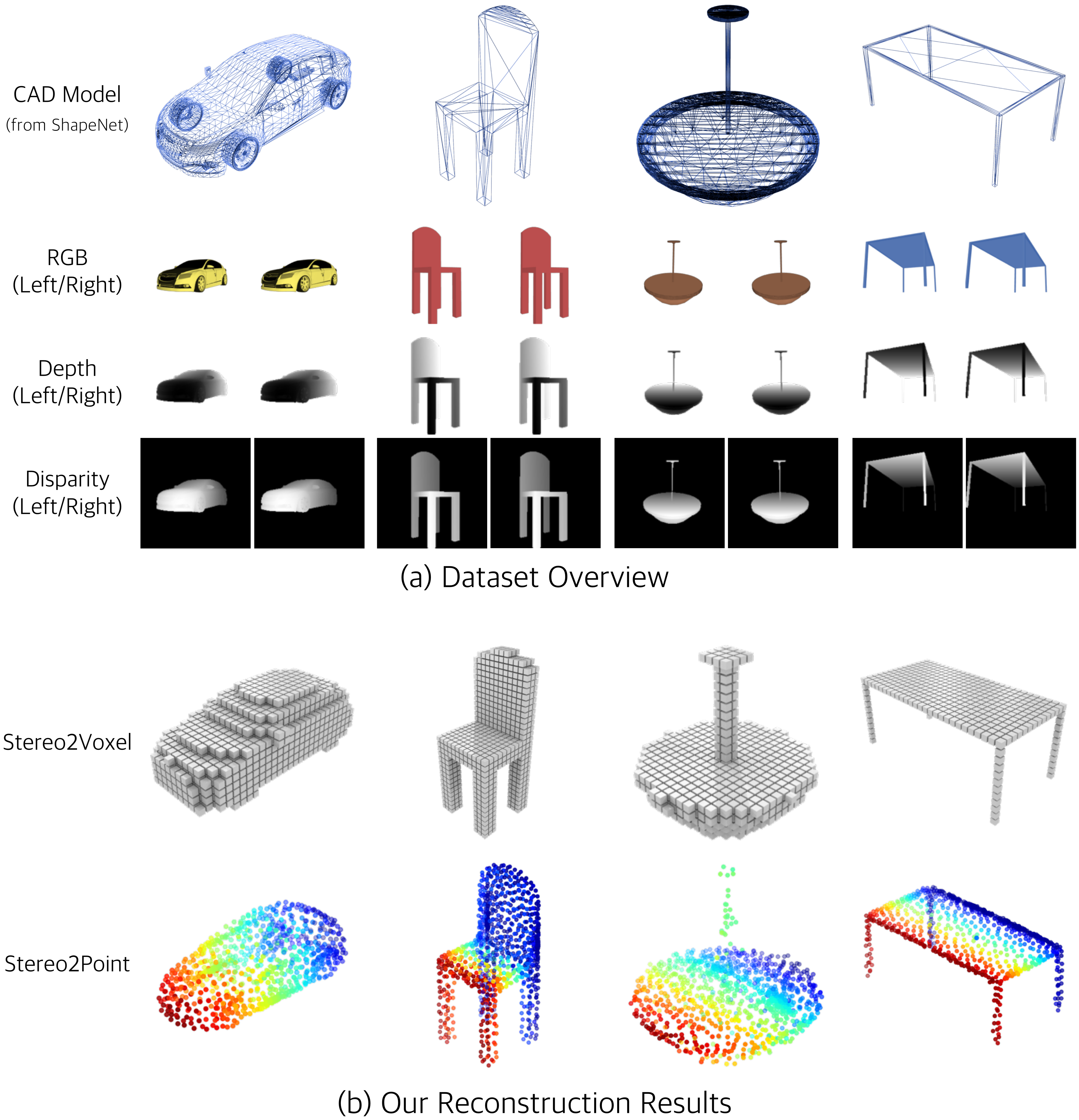}
  }
  \caption{(a) Our dataset provides over 1 million pairs of stereo images along with the corresponding 3D volumes and point clouds, bidirectional depth and disparity maps. (b) We propose two networks to generate a 3D volumes or a point clouds from stereo images, respectively.}
  \label{fig:highlight}
\end{figure}

\begin{figure*}
  \centering
  \resizebox{.8\linewidth}{!} {
    \includegraphics{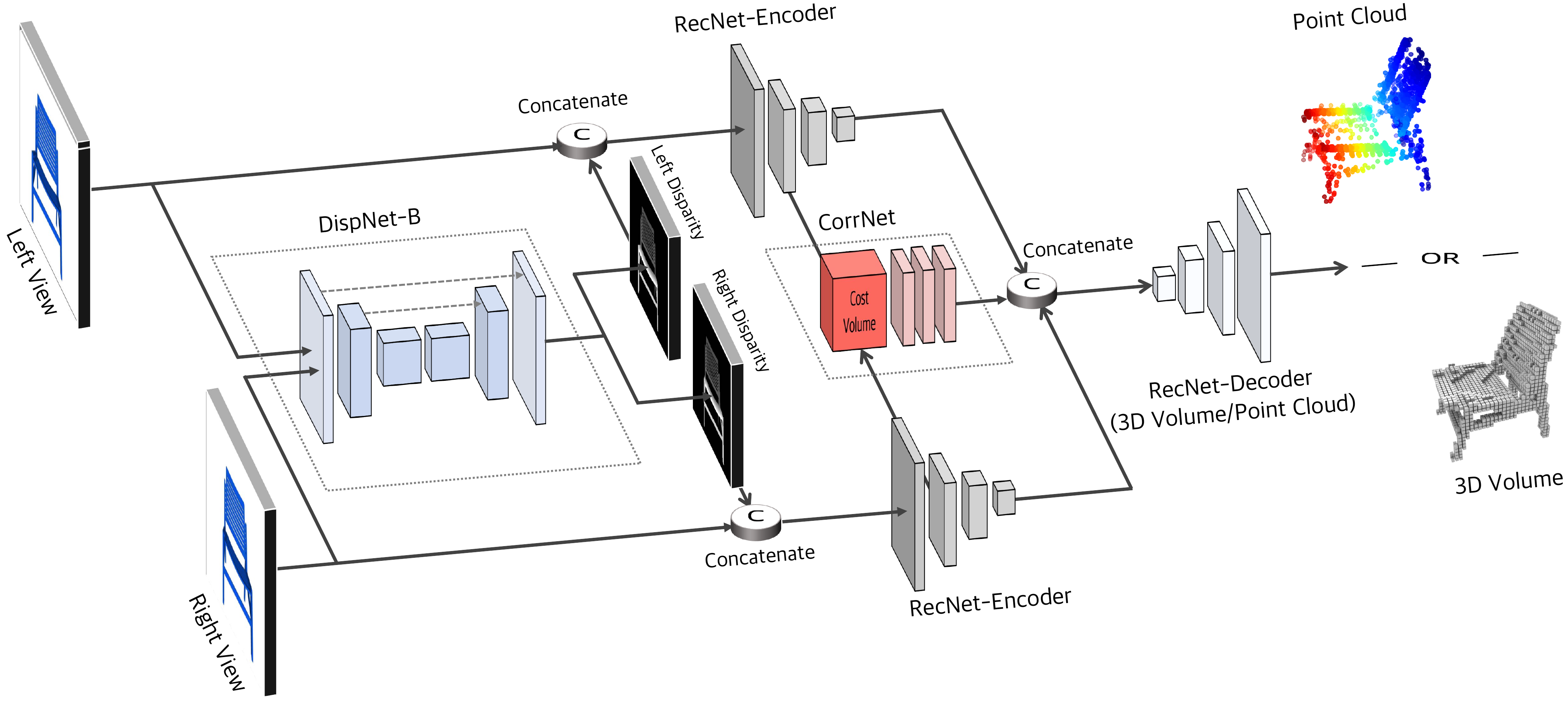}
  }
  \caption{An overview of the proposed methods that recover the 3D volume or point cloud of an object from a pair of stereo images. Note that the weights of the encoders in RecNet are shared between the two views.}
  \label{fig:overview}
\end{figure*}

\section{Introduction}

Recovering the complete 3D shape of an object from a single image is often described as an ill-posed and inherently ambiguous problem because the partial observation of the object can be theoretically associated with an infinite number of possible 3D models.
Data-driven methods \cite{DBLP:conf/iccv/TatarchenkoDB17,DBLP:conf/cvpr/FanSG17} can recover the 3D shape of an object from a single image and have shown impressive results.
However, as demonstrated by \cite{DBLP:conf/cvpr/TatarchenkoRRLKB19}, these methods do not reason about the 3D structure from a single image and rely primarily on recognizing the most similar 3D model from the training set to solve the problem, which may lead to low-quality reconstructions for unseen objects.

It has been proved that depth information benefits object reconstruction \cite{DBLP:conf/nips/0001WXSFT17,DBLP:conf/nips/ZhangZZTF018}.
RGB-D cameras have brought about a profound advancement of object reconstruction because of the use of depth recorded by the depth camera.
However, depth cameras are sensitive to illumination and the depth accuracy degenerates rapidly with the decreasing illumination power \cite{DBLP:journals/arxiv/abs-1812-08125}.
Another way to obtain depth is to estimate depth from an RGB image.
However, monocular depth estimation itself is challenging because of structural variations and object occlusions in most scenes \cite{DBLP:journals/arxiv/abs-1901-09402}.
In contrast, it is more reliable to estimate depth or disparity from stereo images.
In addition, introducing stereo images also enables the networks to explore the dense feature correspondences between two correlated views, which are beneficial to infer the 3D structure of the object.
Despite the above congenital advantages, the surge in dual-lens smartphones and robots also sheds light on 3D object reconstruction from stereo images.

In this paper, we propose a new framework for reconstructing the complete 3D shape of an object from a pair of stereo images.
To explicitly infer the 3D structure of the object, the proposed framework explores the bidirectional disparities and feature correspondences between two views of stereo images.
As illustrated in Figure \ref{fig:overview}, it consists of three sub-networks: DispNet-B, CorrNet, and RecNet.
DispNet-B estimates the bidirectional disparities from a pair of stereo images, which are fed, along with the stereo RGB images, into the encoder of RecNet.
CorrNet finds dense feature correspondences between the stereo image pairs.
The decoder of RecNet generates the 3D volume or point cloud of an object from concatenated feature maps.

Currently, there is no particular dataset for training and benchmarking stereo 3D object reconstruction in the deep learning community.
Therefore, we construct a large-scale synthetic dataset rendered from ShapeNet, named {\it StereoShapeNet}, using the open-source 3D creation suite Blender\footnote{\url{https://www.blender.org}}.
In particular, it includes 1,052,976 pairs of stereo RGB images and the corresponding ground truth for the 3D model, bidirectional depth as well as disparity maps.

The contributions can be summarized as follows:

\begin{itemize}
  \item We study the stereo 3D object reconstruction problem with deep learning. We propose a framework that exploits the disparities and feature correspondences of a pair of stereo images to reconstruct the 3D shape of an object in forms of a 3D volume and point clouds.
  \item We release a large-scale benchmarking dataset, namely {\it StereoShapeNet}, containing over 1M pairs of stereo images rendered from ShapeNet along with the corresponding bidirectional depth and disparity maps, which is the first benchmark for stereo 3D object reconstruction.
  \item Experimental results on the {\it StereoShapeNet} dataset demonstrate that the proposed method outperforms state-of-the-art methods in both point cloud and 3D volume reconstruction.
\end{itemize}

\section{Related Work}

\begin{figure*}
  \centering
  \resizebox{\linewidth}{!} {
    \includegraphics{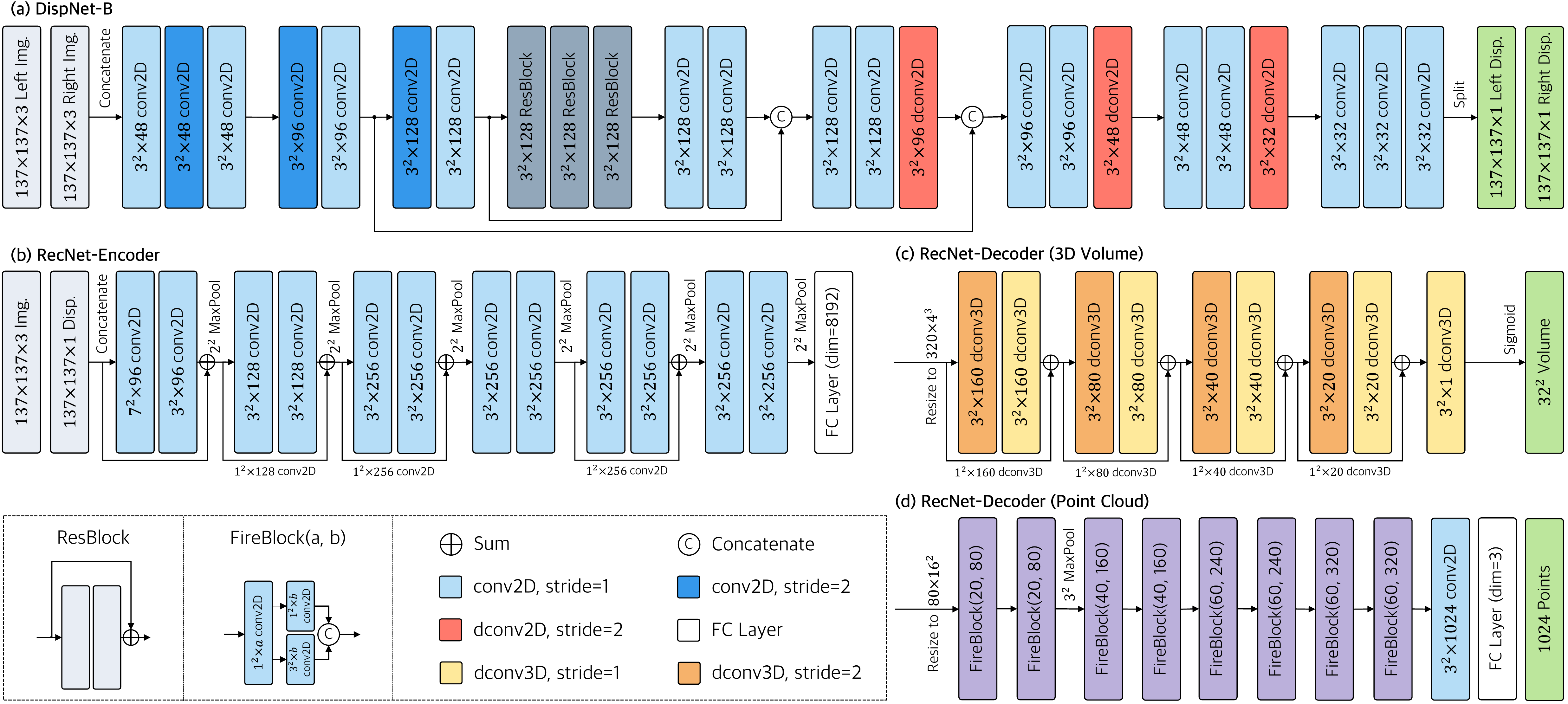}
  }
  \caption{The network architecture of Stereo2Voxel and Stereo2Point. The difference between Stereo2Voxel and Stereo2Point is that Stereo2Voxel uses a RecNet-Decoder (3D Volume) while Stereo2Point adopts a RecNet-Decoder (Point Cloud). The rest of the components are the same in both networks.}
  \label{fig:network-architecture}
\end{figure*}

\noindent \textbf{Single-view 3D Reconstruction.}
Early works such as ShapeFromX \cite{GoogleScholar:aloimonos1988shape,GoogleScholar:buelthoff1991shape} make strong assumptions about the shape or the environment lighting conditions.
Boosted by large-scale datasets of 3D CAD models such as ShapeNet \cite{DBLP:conf/cvpr/WuSKYZTX15}, data-driven single-view 3D reconstruction methods \cite{DBLP:conf/iccv/TatarchenkoDB17,DBLP:conf/cvpr/FanSG17} are able to reconstruct the 3D shape of an object based on only one image without assumptions about the environment.
Tatarchenko et al. \cite{DBLP:conf/iccv/TatarchenkoDB17} propose an octree-based voxel representation and generate a 3D shape from a single image.
PSGN \cite{DBLP:conf/cvpr/FanSG17} recovers the point cloud of an object from a single image.

\noindent \textbf{Multi-view 3D Reconstruction.} 
Classical 3D reconstruction methods, such as SfM and vSLAM, require a collection of RGB images to reconstruct the 3D shape of an object.
The geometric shape is recovered by dense feature extraction and matching \cite{DBLP:conf/iccv/NewcombeLD11}, or by directly minimizing reprojection errors \cite{DBLP:journals/ijcv/BakerM04} from color images. 
Recently, deep neural networks are designed to recover the 3D shape of an object from multiple images  \cite{DBLP:conf/eccv/ChoyXGCS16,DBLP:conf/iccv/XieHXSS19}.
3D-R2N2 \cite{DBLP:conf/eccv/ChoyXGCS16} uses a gated recurrent unit (GRU) \cite{DBLP:journals/arxiv/ChungGCB14} to infer 3D shape from single or multiple images and achieve impressive results.
In Pix2Vox \cite{DBLP:conf/iccv/XieHXSS19}, the context-aware fusion is incorporated to select high-quality reconstructions for each part from different coarse 3D volumes.

\noindent \textbf{RGB-D 3D Reconstruction.}
Several representative works \cite{DBLP:journals/pami/YangRMTW18,DBLP:journals/tvcg/LiSWZ17} reconstruct the 3D shape of an object from RGB-D images by shape completion.
Traditional reconstruction approaches use interpolation techniques ({\it e.g.}, plane fitting, Laplacian hole filling) to infer the  underlying 3D structure \cite{DBLP:conf/graphite/NealenISA06,DBLP:journals/vc/ZhaoGL07}.
3D-RecNet++ \cite{DBLP:journals/pami/YangRMTW18} generates a complete and fine-grained 3D structure from a single depth view with deep neural networks.

\section{The Method}

\subsection{Problem and Notations}

Our goal is to reconstruct the 3D shape of an object from a pair of stereo RGB images. The reconstructed 3D shape is represented in two forms: a 3D volume and a point cloud, as shown in Figure \ref{fig:highlight}.
Assume $I^L$ and $I^R$ indicate the left and right view of an object.
The ground truth disparity maps for the left and right views are denoted as $D^L$ and $D^R$.
The estimated disparity maps are represented by $\hat{D}^L$ and $\hat{D}^R$, respectively.
Let $V$ and $\hat{V}$ be the ground truth and predicted 3D volumes, respectively.
The ground truth and predicted point cloud are denoted by $P = \{(x_i, y_i, z_i)\}_{i=1}^{n_{gt}}$ and $\hat{P} = \{(x_i, y_i, z_i)\}_{i=1}^{n_{p}}$, respectively.

\subsection{Network Architectures}

There are two alternative implementations for generating 3D volumes and point clouds, named {\it Stereo2Voxel} and {\it Stereo2Point}, respectively.
The detailed network architectures are shown in Figure \ref{fig:network-architecture}.
Both of them consist of three sub-networks: DispNet-B, RecNet, and CorrNet.
First, DispNet-B predicts bidirectional disparities from a pair of stereo images.
Second, the encoder of RecNet generates two feature vectors of size $8192$ from stereo images and their disparities.
Third, the feature maps from the 3rd convolutional layer of the encoder in RecNet are forwarded to CorrNet, which finds the feature correspondences between the two images and produces a feature vector of size $4096$.
Finally, the decoder of RecNet reconstructs a 3D volume or a point cloud from the concatenated feature vectors generated from the encoder of RecNet and CorrNet.
In RecNet, the decoders for {\it Stereo2Voxel} and {\it Stereo2Point} are different.
{\it Stereo2Voxel} outputs a 3D volume with a resolution of $32^3$ while {\it Stereo2Point} outputs an unordered point set containing 1024 points in the object's canonical view.

\noindent \textbf{DispNet-B.}
DispNet-B is to compute the bidirectional disparities.
The U-Net based structure of DispNet-B is shown in Figure \ref{fig:network-architecture} (a).
The encoder outputs feature maps with $\frac{1}{8} \times \frac{1}{8}$ of the input size.
Afterward, the following decoder produces the full resolution disparities by three transposed convolutional layers.
Different from DispNet \cite{DBLP:conf/cvpr/MayerIHFCDB16}, the proposed DispNet-B predicts bidirectional disparities in only one forward computation.
It has been proved that bidirectional prediction is better than unidirectional prediction in optical flow estimation \cite{DBLP:conf/eccv/IlgSKB18}.
DispNet-B is 6\% size of DispNet because it takes a small number of channels for each layer.
Therefore, DispNet-B is computationally efficient and 4 times faster than DispNet.
The output of DispNet-B is bidirectional disparities with the same size as the inputs.

\noindent \textbf{RecNet.}
RecNet produces 3D volumes and point clouds by taking stereo RGB images and the corresponding disparities as input.
Inspired by the fact that residual connections between convolutional layers accelerate the optimization process \cite{DBLP:conf/cvpr/HeZRS16},  the encoder of RecNet uses the residual block as the building block.
To match the number of channels after convolutions, we add a $1 \times 1$ convolutional layer for residual connections.
There are two versions of decoders to transform feature vectors into 3D volumes ({\it Stereo2Voxel}) and point clouds ({\it Stereo2Point}), respectively, as shown in Figure \ref{fig:network-architecture} (c) and (d).
The input feature vectors of the decoders are concatenated of the feature vectors from the encoder of RecNet and CorrNet.

To generate 3D volumes, the decoder of RecNet contains nine transposed 3D convolutional layers that upsample the feature maps to the size of $32^3$.
The final feature map is passed to a sigmoid layer and produces the probabilities of each 3D occupancy grid.
Similar to the encoder of RecNet, there are residual connections between the transposed 3D convolutional layers for improving efficiency.

To produce point clouds, the decoder consists of eight Fire modules \cite{DBLP:conf/iclr/IandolaMAHDK16} and a fully connected layer.
Following PSGN \cite{DBLP:conf/cvpr/FanSG17}, we produce a matrix of size $1024 \times 3$, which corresponds to the coordinates of $1024$ points.
To reduce the size of the decoder, we adopt several Fire modules to replace the massive transposed convolutional layers and fully connected layers.
Each Fire block contains a squeeze convolutional layer (which has only $1 \times 1$ filters), fed into an expand layer that has a mix of $1 \times 1$ and $3 \times 3$ convolution filters.
Fire blocks replace several $3 \times 3$ filters with $1 \times 1$ filters and decrease the number of input channels to $3 \times 3$ filters.
Consequently, the RecNet is 28\% size of PSGN.

\noindent \textbf{CorrNet.}
As stated above, CorrNet is introduced to find the feature correspondences between the two views.
The feature correspondences are computed by forming a cost volume that preserves the knowledge of geometry of stereo vision.
Following GC-Net \cite{DBLP:conf/iccv/KendallMDH17}, the cost volume is of dimensionality {\it height}$\times${\it width}$\times${\it (max disparity / shift intervals)}$\times${\it channels} and formed by stacking left and right feature maps with shift intervals.
The cost volume allows the network to perform multiplicative patch comparisons between two feature maps by aggregating feature information along the disparity dimension as well as spatial dimensions.

We propose to use a 3D-CNN for further feature matching.
In 3D-CNN, there are nine sets of 3D convolutional layers with output channels of $128$, along with the corresponding batch normalization layers and ReLU layers.
The convolutional layers are with kernel sizes of $3 \times 3 \times 3$ except for the first layer with a kernel size of $1 \times 1 \times 1$.
3D-CNN is lastly followed by a $1 \times 1 \times 1$ convolutional layer with an output channel of $1$ and a $1 \times 1$ convolutional layer whose output channel is $1$.
Both of them are followed by a batch normalization layer and a ReLU activation.
The output of 3D-CNN is then flattened and passed to a fully connected layer with a dimension of $4096$.

\subsection{Loss Functions}

There are three loss functions for predicting disparity maps, 3D volumes, and point clouds, respectively.

\noindent \textbf{Disparity Loss.}
We adopt Mean Square Error (MSE) loss to measure the difference between the estimated disparities and the corresponding ground truth.
More specifically, the disparity loss is defined as

\begin{equation}
  \mathcal{L}_{disp}
  = \frac{1}{HW} \sum \left[||\hat{D}^L - D^L||^2 
  + ||\hat{D}^R - D^R||^2\right]
  \label{eq:mse-loss}
\end{equation}
where $H$ and $W$ represent the height and width of disparity, respectively.

\noindent \textbf{3D Volume Loss.}
The loss function for predicting 3D volume is defined as the sum value of the voxel-wise binary cross entropies between the reconstructed object and the ground truth.
More formally, it can be defined as

\begin{equation}
  \mathcal{L}_{vol} 
  = \frac{1}{n_{vox}} \sum \left[V \log(\hat{V}) + (1 - V) \log(1 - \hat{V})\right]
  \label{eq:bce-loss}
\end{equation}
where $n_{vox}$ denotes the number of voxels in the 3D volume.

\noindent \textbf{Point Cloud Loss.}
We use the Chamfer Distance to measure the point-to-point similarity between two point clouds following \cite{DBLP:conf/cvpr/FanSG17}.
The Chamfer Distance (CD) is defined as

\begin{equation}
  \mathcal{L}_{cd} 
  = \frac{1}{n_{gt}} \sum_{p \in P} \min_{q \in \hat{P}} ||p - q||^2_2
  + \frac{1}{n_{p}} \sum_{p \in \hat{P}} \min_{q \in P} ||p - q||^2_2
  \label{eq:cd}
\end{equation}

\section{Experiments}

\begin{table*}[!t]
  \centering
  \caption{Reconstruction results from stereo RGB images on {\it StereoShapeNet}. Intersection over union (IoU) and Chamfer distance (CD) are adopted to evaluate the performance for voxel reconstruction and point cloud reconstruction, respectively. The best number for each category is highlighted in bold. Note that PSGN and PSGN$^*$ take object masks as an additional input. LSM takes extrinsic camera parameters as an additional input.}
  \resizebox{\linewidth}{!} {
    \begin{tabular}{l|ccccc|ccccc}
    \toprule
      \multirow{2}{*}{Category} &
      \multicolumn{5}{c|}{Voxel Reconstruction (IoU, Resolutions of $32^3$)} &
      \multicolumn{5}{c}{Point Cloud Reconstruction (CD $\times 10^{-3}$, 1024 Points)} \\
      \noalign{\smallskip} \cline{2-11} \noalign{\smallskip} &
                  Matryoshka   & Matryoshka$^*$ & Pix2Vox  & LSM   &
                  Stereo2Voxel & 
                  PSGN         & PSGN$^*$ & AtlasNet & AtlasNet$^*$ &
                  Stereo2Point \\
      \midrule
      \midrule
      airplane   & 0.557       & 0.535    & 0.686    & 0.621
                 & \bf{0.709} 
                 & 0.826       & 0.699    & 0.807    & 0.796
                 & \bf{0.534} \\
      bench      & 0.524       & 0.473    & 0.566    & 0.517
                 & \bf{0.622} 
                 & 1.789       & 1.695    & 1.996    & 1.796
                 & \bf{1.182} \\
      cabinet    & 0.766       & 0.763    & 0.754    & 0.691
                 & \bf{0.784} 
                 & 2.360       & 1.853    & 1.756    & 1.692
                 & \bf{1.229} \\
      car        & 0.827       & 0.810    & 0.811    & 0.796
                 & \bf{0.830}
                 & 1.295       & 0.882    & 1.045    & 1.036
                 & \bf{0.779} \\
      chair      & 0.559       & 0.514    & 0.604    & 0.595
                 & \bf{0.669} 
                 & 2.004       & 1.594    & 1.837    & 1.858
                 & \bf{1.267} \\
      display    & 0.635       & 0.614    & 0.586    & 0.547
                 & \bf{0.692} 
                 & 2.815       & 2.238    & 2.386    & 2.146
                 & \bf{1.356} \\
      lamp       & 0.424       & 0.411    & 0.449    & 0.469
                 & \bf{0.521} 
                 & 3.973       & 3.038    & 4.142    & 4.118
                 & \bf{3.001} \\
      speaker    & 0.697       & 0.727    & 0.658    & 0.670
                 & \bf{0.701}
                 & 3.868       & 2.691    & 2.839    & 2.869
                 & \bf{2.124} \\
      rifle      & 0.540       & 0.557    & 0.652    & 0.682
                 & \bf{0.690}
                 & 0.790       & 0.763    & 0.818    & 0.874
                 & \bf{0.524} \\
      sofa       & 0.702       & 0.679    & 0.714    & 0.651
                 & \bf{0.770} 
                 & 2.625       & 2.086    & 1.664    & 1.656
                 & \bf{1.199} \\
      table      & 0.559       & 0.503    & 0.570    & 0.566
                 & \bf{0.635}
                 & 1.889       & 1.500    & 1.892    & 1.916
                 & \bf{1.337} \\
      telephone  & 0.759       & 0.847    & 0.831    & 0.694
                 & \bf{0.866} 
                 & 1.445       & 1.158    & 1.156    & 1.250
                 & \bf{0.896} \\
      watercraft & 0.587       & 0.595    & 0.558    & 0.592
                 & \bf{0.645} 
                 & 2.029       & 1.495    & 1.712    & 1.524
                 & \bf{1.027} \\
      \midrule
      Overall    & 0.626       & 0.603    & 0.652    & 0.632
                 & \bf{0.702} 
                 & 1.916       & 1.493    & 1.704    & 1.689
                 & \bf{1.185} \\
      \bottomrule
    \end{tabular}
  }
  \label{tab:stereo-rgb-reconstruction}
\end{table*}

\begin{figure*}[!t]
  \centering
  \resizebox{\linewidth}{!} {
    \includegraphics{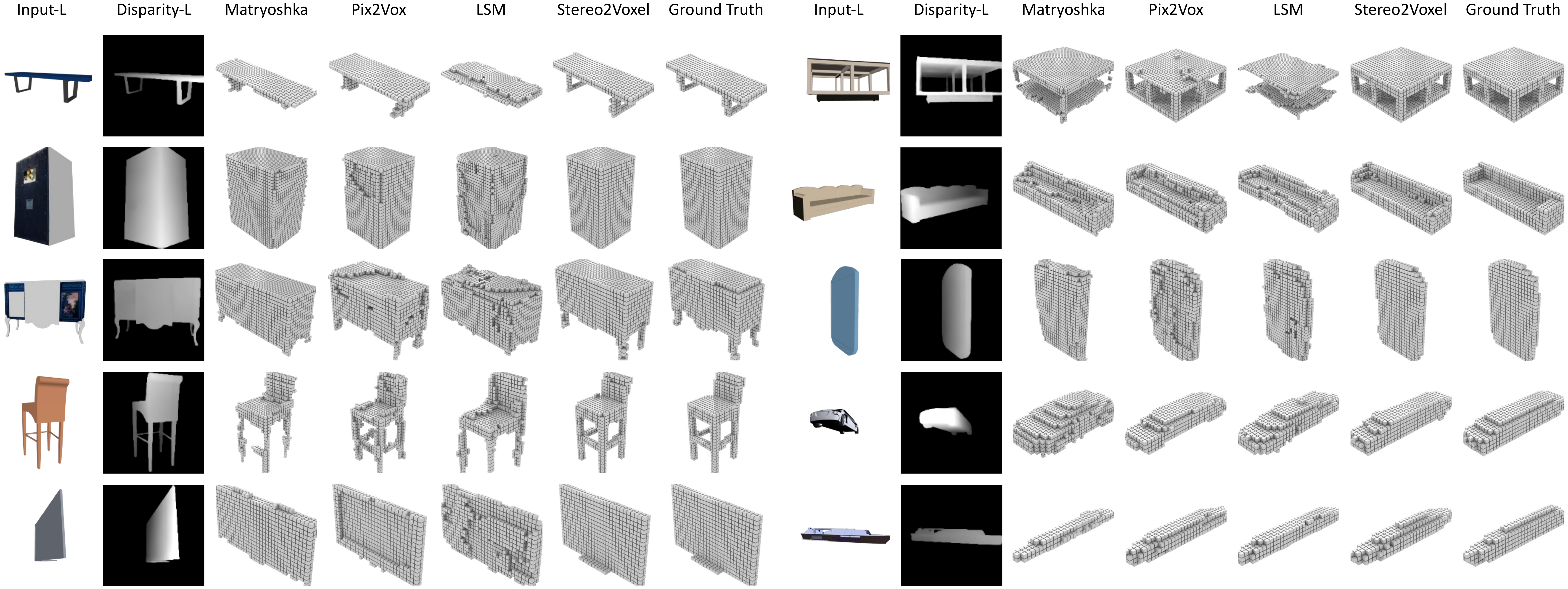}
  }
  \caption{The 3D volumes reconstructed from stereo images on the {\it StereoShapeNet} dataset. Input-L represents the left-view of the input images. Disparity-L denotes the left-view of the estimated disparity by DispNet-B.}
  \label{fig:volume-reconstruction}
\end{figure*}

\begin{figure*}[!t]
  \centering
  \resizebox{\linewidth}{!} {
    \includegraphics{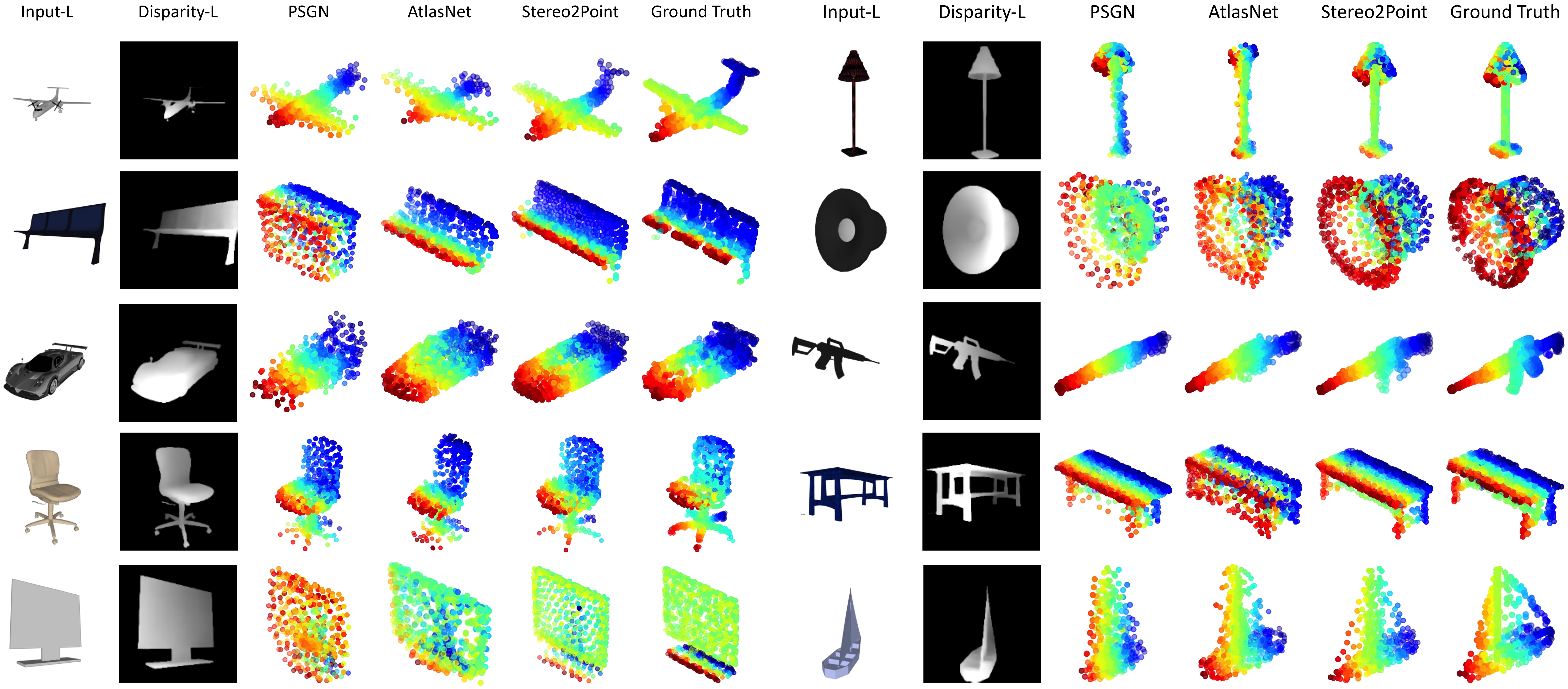}
  }
  \caption{The point cloud reconstructed from stereo images on the {\it StereoShapeNet} dataset. Input-L represents the left-view of the input imag`33es. Disparity-L denotes the left-view of the estimated disparity by DispNet-B. Note that PSGN takes object masks as an additional input.}
  \label{fig:point-reconstruction}
\end{figure*}

\subsection{Dataset and Metrics}

\noindent \textbf{Dataset.}
We present a large-scale synthetic dataset, named {\it StereoShapeNet}.
We use the 3D models in the ShapeNet dataset \cite{DBLP:conf/cvpr/WuSKYZTX15} to generate stereo images, depth maps, and disparity maps for all our experiments.
Specifically, we use a subset of ShapeNet consisting of 44k 3D models from 13 major categories following the settings of 3D-R2N2 \cite{DBLP:conf/eccv/ChoyXGCS16}.
We generate a large set of stereo renderings for the models sampled from a viewing sphere $\theta_{az} \in [0, 360)$ and $\theta_{el} \in [-20, 30]$ degrees with the open-source 3D computer graphics toolset Blender.
We also render the depth images as well as disparities corresponding to each rendered image.
The images are with resolutions of $224 \times 224$.
In the Blender, the stereo camera is of a virtual focal length of 35mm and a simulated sensor of size 32mm wide.
The baseline of the stereo camera is 130mm.
In total, the {\it StereoShapeNet} contains 1,052,976 pairs of stereo images with the corresponding bidirectional depth and disparity maps. 
For volumetric ground truth, we voxelize each 3D CAD model at a resolution of $32^3$.
For the ground truth of point clouds, we uniformly densify 3D points on the surfaces to generate 16,384 points for each 3D model.

\noindent \textbf{Evaluation Metrics.}
To evaluate the quality of the output from {\it Stereo2Voxel}, we binarize the probabilities at a fixed threshold of $0.4$ and use intersection over union (IoU) as the similarity measure.
More specifically,

\begin{equation}
  {\rm IoU} = \frac{\sum_{i, j, k} {\rm I}(\hat{V}^{(i, j, k)} > t) {\rm I}(V^{(i, j, k)})}{\sum_{i, j, k} {\rm I}\left[ {\rm I}(\hat{V}^{(i, j, k)} > t) + {\rm I}(V^{(i, j, k)}) \right]}
\end{equation}
where $\hat{V}^{(i, j, k)}$ and $V^{(i, j, k)}$ represent the predicted occupancy probability and the ground truth at $(i, j, k)$, respectively.
${\rm I}(\cdot)$ is an indicator function and $t$ denotes a voxelization threshold.
Higher IoU values indicate better reconstruction results.

To evaluate the similarity between the ground truth and the generated point clouds from {\it Stereo2Point}, we introduce Chamfer Distance (CD) following \cite{DBLP:conf/cvpr/FanSG17} (Equation \ref{eq:cd}).
The lower the CD value is, the closer the prediction is to the ground truth.

\subsection{Implementation Details}

We use $137 \times 137$ RGB images as input to train the proposed methods with a batch of $20$.
The 3D volume generated by {\it Stereo2Voxel} is $32^3$ in size.
Following \cite{DBLP:conf/cvpr/FanSG17}, $n_p$ and $n_{gt}$ are pre-assigned to $1024$ and $16384$, respectively.
The shift interval in CorrNet is set to $1$.
We implement our network in PyTorch\footnote{\url{https://github.com/hzxie/Stereo-3D-Reconstruction}} using an Adam \cite{DBLP:journals/arxiv/KingmaB14} optimizer with $\beta_1 = 0.9$ and $\beta_2 = 0.999$.
The initial learning rate is set to $10^{-4}$ and decayed by 2 after 300 epochs.
The optimization is set to stop after 500 epochs.

\subsection{Reconstruction results from Synthetic Images}

To evaluate the performance of the proposed methods in handling synthetic images, we compare our methods against several state-of-the-art methods on the {\it StereoShapeNet} testing set.
We fine-tune all competitive methods on the {\it StereoShapeNet} dataset and test them with the same pair of stereo images for each object.
For voxel reconstruction methods, we compare {\it Stereo2Voxel} with Matryoshka Network \cite{DBLP:conf/cvpr/Richter018} and Pix2Vox \cite{DBLP:conf/iccv/XieHXSS19} that are used for single-view and multi-view 3D object reconstruction, respectively.
To further demonstrate the superior reconstruction ability of the proposed methods, we compare {\it Stereo2Voxel} with a MVS method LSM \cite{DBLP:conf/nips/KarHM17}.
For point cloud reconstruction, we compare {\it Stereo2Point} with PSGN \cite{DBLP:conf/cvpr/FanSG17} and AtlasNet \cite{DBLP:conf/cvpr/GroueixFKRA18}.
For single-view reconstruction methods, the left view of an object is fed into the networks.
To make a fair comparison with single-view reconstruction methods, we extend these methods by taking the concatenation of two stereo images, denoted as Matryoshka$^*$, PSGN$^*$, and AtlasNet$^*$, respectively.

Table \ref{tab:stereo-rgb-reconstruction} shows the accuracy of reconstruction results from a pair of stereo images for 13 major categories on {\it StereoShapeNet}.
Experimental results indicate that both {\it Stereo2Voxel} and {\it Stereo2Points} outperform state-of-the-art methods for single-view and multi-view reconstruction.
Moreover, compared with MVS methods, {\it Stereo2Voxel} outperforms LSM, which takes extrinsic camera parameters as an additional input.
Figures \ref{fig:volume-reconstruction} and \ref{fig:point-reconstruction} show several reconstruction examples on the {\it StereoShapeNet} testing set.
Both {\it Stereo2Voxel} and {\it Stereo2Point} recover better details of objects ({\it e.g.}, table legs and chair legs) compared to the state-of-the-art methods.

\subsection{Reconstruction Results from Naturalistic Images}

\begin{figure}[!t]
  \resizebox{\linewidth}{!} {
    \includegraphics{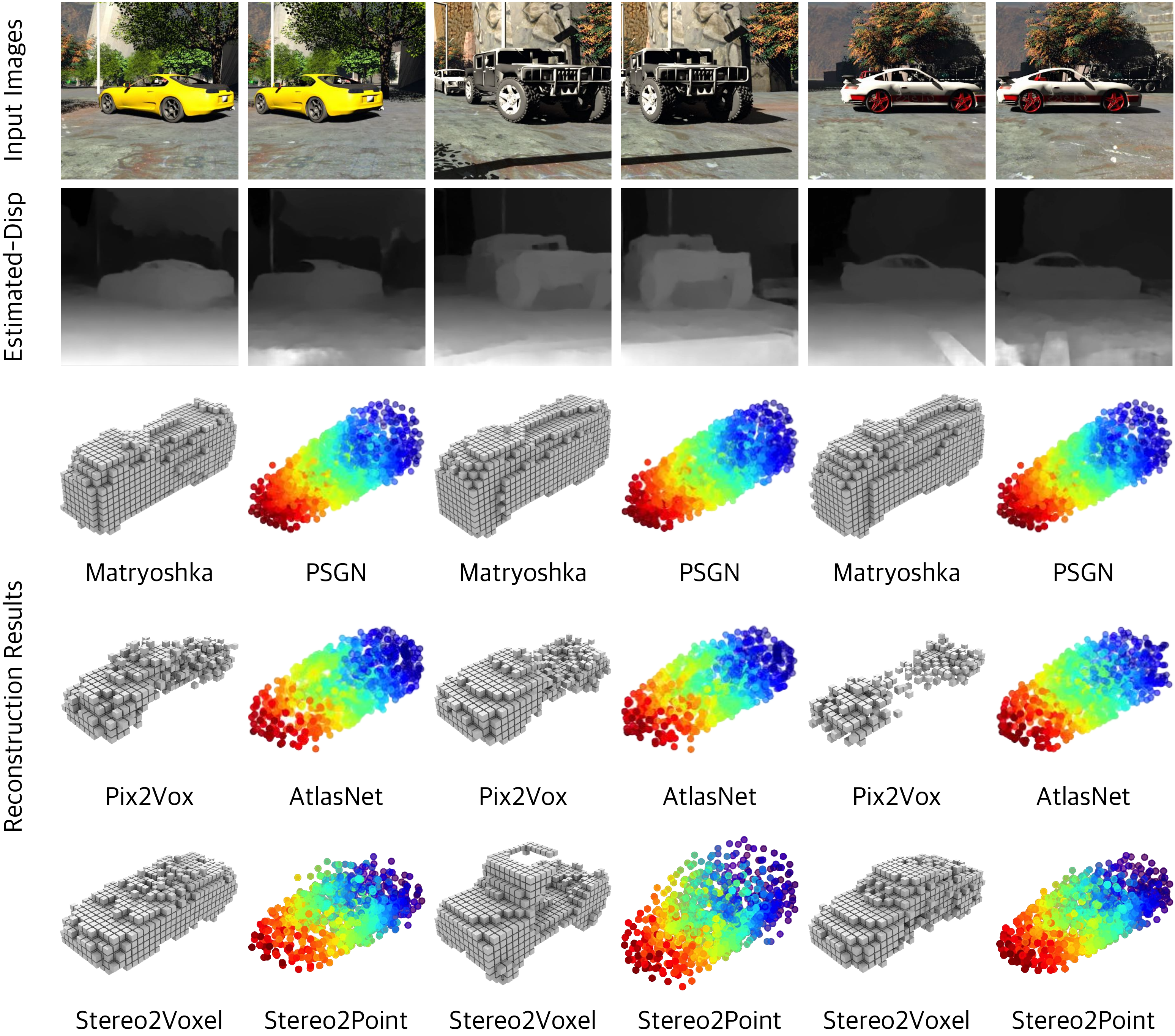}
  }
  \caption{Reconstruction results compared to the state-of-the-arts methods on the Driving dataset. Estimated-Disp represents the disparities estimated by DispNet-B. Note that PSGN takes object masks as an additional input.}
  \label{fig:real-world-reconstruction}
\end{figure}

To evaluate the performance of the proposed methods in handling naturalistic images, we compare our methods against Matryoshka, Pix2Vox, PSGN, and AltasNet on a subset of the Driving dataset \cite{DBLP:conf/cvpr/MayerIHFCDB16}. 
We fine-tune all methods on the car category of the {\it StereoShapeNet} dataset using backgrounds that are randomly sampled from the SUN database \cite{DBLP:conf/cvpr/XiaoHEOT10}.
We further augment our training data by random color and light jittering.
For better generalize to naturalistic scenarios, we pretrain DispNet-B on the FlyingThings 3D dataset \cite{DBLP:conf/cvpr/MayerIHFCDB16}. 

For testing, we select several untruncated and unoccluded car images in the Driving dataset.
First, the images are cropped according to the bounding box of the largest cars within the image.
Then, these cropped images are rescaled to the input size of the networks.
Figure \ref{fig:real-world-reconstruction} shows some representative reconstruction results of {\it Stereo2Voxel} and {\it Stereo2Point} compared with other methods on the Driving dataset.
Except for Pix2Vox, all competitive methods produce almost the same reconstruction results for the three images, which indicates that single-view reconstruction methods rarely reason about the 3D structure of an object and tend to output a mean shape to minimize the reconstruction errors.
In contrast, {\it Stereo2Voxel} and {\it Stereo2Point} recovers better skeletons of  objects than other methods.

\subsection{Ablation Study}

\begin{table}[!t]
  \centering
  \caption{The ablation studies of DispNet-B and CorrNet. Intersection over union (IoU) and Chamfer distance (CD) are adopted to evaluate the performance for {\it Stereo2Voxel} and {\it Stereo2Point}, respectively. Note that the CD is computed on 1024 points and multiplied by $10^3$.}
  \begin{tabularx}{\linewidth}{cY||c|c}
  	\toprule
  	DispNet-B  & CorrNet    & Stereo2Voxel & Stereo2Point \\
  	\midrule
  	\checkmark & \checkmark & \bf{0.702}   & \bf{1.185} \\
  	\checkmark & $\times$   & 0.690        & 1.284 \\
  	$\times$   & \checkmark & 0.678        & 1.379 \\
  	$\times$   & $\times$   & 0.651        & 1.570 \\
  	\bottomrule
  \end{tabularx}
  \label{tab:ablation-study-dispnet-corrnet}
\end{table}

\begin{table}[!t]
  \centering
  \caption{The comparison of the numbers of parameters, inference times, and the endpoint error (EPE) on the StereoShapeNet and the subset of FlyingThings 3D and dataset. Note that the inference time is for a 960$\times$540 image on an NVIDIA GTX 1080 Ti GPU.}
  \resizebox{\linewidth}{!} {
    \begin{tabular}{l|ccc}
      \toprule
      Methods               & DispNet-B       & DispNet & GA-Net \\
      \midrule
      \#Parameters (M)      & \bf{2.54}       & 39.45   & 6.58 \\
      Inference Time (s)    & \bf{0.018}/pair & 0.063   & 6.999 \\
      FlyingThings 3D (EPE) & 1.292           & 1.157   & \bf{0.515} \\
      StereoShapeNet (EPE)  & 0.096           & 0.092   & \bf{0.089} \\
      \bottomrule
    \end{tabular}
  }
  \label{tab:DispNet-B-comparison}
\end{table}

\begin{table}[!t]
  \centering
  \caption{The comparison of reconstruction results with different disparity map quality on the StereoShapeNet dataset. Intersection over union (IoU) and Chamfer distance (CD) are adopted to evaluate the performance for {\it Stereo2Voxel} and {\it Stereo2Point}, respectively. Note that the CD is computed on 1024 points and multiplied by $10^3$.}
  \begin{tabularx}{\linewidth}{X|Y|Y}
  \toprule
  Methods      & Stereo2Voxel & Stereo2Point \\
  \midrule
  SGBM         & 0.680        & 1.282 \\
  DispNet-B    & 0.702        & 1.185 \\
  DispNet      & 0.702        & 1.184 \\
  GA-Net       & \bf{0.705}   & \bf{1.178} \\
  \bottomrule
  \end{tabularx}
  \label{tab:disparities-reconstruction-comparison}
\end{table}

In this section, we validate the effectiveness of two key components of our method: DispNet-B and CorrNet.

\noindent \textbf{DispNet-B.}
Both {\it Stereo2Voxel} and {\it Stereo2Point} estimate the disparity maps from a pair of stereo images.
To demonstrate the importance of disparities in stereo 3D reconstruction, we remove DispNet-B from the proposed methods.
The stereo RGB images are directly fed into the encoders in RecNet without estimating disparities.
As illustrated in Table \ref{tab:ablation-study-dispnet-corrnet}, removing DispNet-B causes an increase of $16.4\%$ in CD for {\it Stereo2Points}.
The IoU decreases to $0.678$ when removing DispNet-B from {\it Stereo2Voxel}.

\noindent \textbf{CorrNet.}
CorrNet aims to find feature correspondences between a pair of stereo images.
To quantitatively evaluate CorrNet, we compare the performance of {\it Stereo2Voxel} and {\it Stereo2Point} without CorrNet.
As shown in Table \ref{tab:ablation-study-dispnet-corrnet}, the CD increases to $1.284$ when CorrNet is removed from {\it Stereo2Points}.
The IoU decreases to $0.690$ when removing CorrNet from {\it Stereo2Voxel}.
Moreover, removing both DispNet-B and CorrNet results in worse reconstruction results in both {\it Stereo2Point} and {\it Stereo2Voxel}, where the CD increases by $32.5\%$ in {\it Stereo2Point} and the IoU decreases by $7.3\%$ in {\it Stereo2Voxel}, respectively.

\subsection{Discussion}

\noindent \textbf{Performance Evaluation of DispNet-B.}
To further compare DispNet-B with other stereo matching methods, we evaluate the performance of DispNet-B on the subset of Flying Things 3D (clean pass, disparity $<$ 96 pixels) test dataset.
 Since we only care about the results of predicted disparity in non-occluded regions, we adopt the endpoint error (EPE) on non-occluded regions as the measure.
 As shown in Table \ref{tab:DispNet-B-comparison}, the EPE of DispNet-B is comparable with DispNet \cite{DBLP:conf/cvpr/MayerIHFCDB16} and worse than GA-Net \cite{DBLP:conf/cvpr/ZhangVRP19}.
However, DispNet-B is only 6\% size of DispNet and 38\% size of GA-Net.
In terms of inferring time, DispNet-B is about 7 and 778 times faster than DispNet and GA-Net, respectively.
Moreover, DispNet-B can predict the bidirectional disparity maps for both views simultaneously. 

\noindent \textbf{Effectiveness of Disparity Map Quality.}
To quantitatively compare the reconstruction results of {\it Stereo2Voxel} and {\it Stereo2Points} with different disparities, we replace DispNet-B with SGBM \cite{DBLP:journals/pami/Heiko07}, DispNet, and GA-Net.
As shown in Table \ref{tab:disparities-reconstruction-comparison}, the reconstruction results with disparities estimated by DispNet and GA-Net are slightly better than with disparities estimated by DispNet-B, while SGBM degenerates the reconstruction results.
The experimental results indicate that better disparities lead to better reconstruction results.

\section{Conclusion}

In this paper, we present a novel framework to recover the 3D shape of an object from a pair of stereo images.
The proposed method reasons about the 3D structure by exploring bidirectional disparities and feature corresponding between the two views.
To our best knowledge, our work is the first to study 3D reconstruction from stereo images with deep learning.
In order to support this work and inspire more studies towards this new direction, we also construct a large-scale synthetic dataset, named {\it StereoShapeNet}, which contains 1M pairs of stereo images rendered from ShapeNet along with the corresponding bidirectional depth and disparity maps.
Quantitative and qualitative evaluation for both 3D volumes and point clouds on {\it StereoShapeNet} indicate that the proposed method outperforms state-of-the-art methods.

\noindent \textbf{Acknowledgements}
This work was supported by the National Natural Science Foundation of China under Project No. 61772158, 61702136, 61872112 and U1711265.

\bibliographystyle{aaai}
\bibliography{references}

\begin{thebibliography}{}

\bibitem[\protect\citeauthoryear{Aloimonos}{1988}]{GoogleScholar:aloimonos1988shape}
Aloimonos, J.
\newblock 1988.
\newblock Shape from texture.
\newblock {\em Biological cybernetics} 58(5):345--360.

\bibitem[\protect\citeauthoryear{Baker and
  Matthews}{2004}]{DBLP:journals/ijcv/BakerM04}
Baker, S., and Matthews, I.~A.
\newblock 2004.
\newblock Lucas-kanade 20 years on: {A} unifying framework.
\newblock {\em International Journal of Computer Vision} 56(3):221--255.

\bibitem[\protect\citeauthoryear{Bhoi}{2019}]{DBLP:journals/arxiv/abs-1901-09402}
Bhoi, A.
\newblock 2019.
\newblock Monocular depth estimation: {A} survey.
\newblock {\em arXiv} 1901.09402.

\bibitem[\protect\citeauthoryear{Buelthoff and
  Yuille}{1991}]{GoogleScholar:buelthoff1991shape}
Buelthoff, H.~H., and Yuille, A.~L.
\newblock 1991.
\newblock Shape-from-x: Psychophysics and computation.
\newblock In {\em Sensor Fusion III: 3{D} Perception and Recognition}, volume
  1383,  235--247.
\newblock International Society for Optics and Photonics.

\bibitem[\protect\citeauthoryear{Chen \bgroup et al\mbox.\egroup
  }{2018}]{DBLP:journals/arxiv/abs-1812-08125}
Chen, Y.; Ren, J.; Cheng, X.; Qian, K.; and Gu, J.
\newblock 2018.
\newblock Very power efficient neural time-of-flight.
\newblock {\em arXiv} 1812.08125.

\bibitem[\protect\citeauthoryear{Choy \bgroup et al\mbox.\egroup
  }{2016}]{DBLP:conf/eccv/ChoyXGCS16}
Choy, C.~B.; Xu, D.; Gwak, J.; Chen, K.; and Savarese, S.
\newblock 2016.
\newblock {3D-R2N2}: A unified approach for single and multi-view 3{D} object
  reconstruction.
\newblock In {\em {ECCV} 2016}.

\bibitem[\protect\citeauthoryear{Chung \bgroup et al\mbox.\egroup
  }{2014}]{DBLP:journals/arxiv/ChungGCB14}
Chung, J.; G{\"{u}}l{\c{c}}ehre, {\c{C}}.; Cho, K.; and Bengio, Y.
\newblock 2014.
\newblock Empirical evaluation of gated recurrent neural networks on sequence
  modeling.
\newblock In {\em {NIPS} Workshops 2014}.

\bibitem[\protect\citeauthoryear{Fan, Su, and
  Guibas}{2017}]{DBLP:conf/cvpr/FanSG17}
Fan, H.; Su, H.; and Guibas, L.~J.
\newblock 2017.
\newblock A point set generation network for 3{D} object reconstruction from a
  single image.
\newblock In {\em {CVPR} 2017}.

\bibitem[\protect\citeauthoryear{Groueix \bgroup et al\mbox.\egroup
  }{2018}]{DBLP:conf/cvpr/GroueixFKRA18}
Groueix, T.; Fisher, M.; Kim, V.~G.; Russell, B.~C.; and Aubry, M.
\newblock 2018.
\newblock A papier-m{\^{a}}ch{\'{e}} approach to learning 3{D} surface
  generation.
\newblock In {\em {CVPR} 2018}.

\bibitem[\protect\citeauthoryear{He \bgroup et al\mbox.\egroup
  }{2016}]{DBLP:conf/cvpr/HeZRS16}
He, K.; Zhang, X.; Ren, S.; and Sun, J.
\newblock 2016.
\newblock Deep residual learning for image recognition.
\newblock In {\em {CVPR} 2016}.

\bibitem[\protect\citeauthoryear{Hirschmuller}{2007}]{DBLP:journals/pami/Heiko07}
Hirschmuller, H.
\newblock 2007.
\newblock Stereo processing by semiglobal matching and mutual information.
\newblock {\em TPAMI} 30(2):328--341.

\bibitem[\protect\citeauthoryear{Iandola \bgroup et al\mbox.\egroup
  }{2017}]{DBLP:conf/iclr/IandolaMAHDK16}
Iandola, F.~N.; Moskewicz, M.~W.; Ashraf, K.; Han, S.; Dally, W.~J.; and
  Keutzer, K.
\newblock 2017.
\newblock Squeezenet: Alexnet-level accuracy with 50x fewer parameters and
  {\textless}1mb model size.
\newblock In {\em {ICLR} 2017}.

\bibitem[\protect\citeauthoryear{Ilg \bgroup et al\mbox.\egroup
  }{2018}]{DBLP:conf/eccv/IlgSKB18}
Ilg, E.; Saikia, T.; Keuper, M.; and Brox, T.
\newblock 2018.
\newblock Occlusions, motion and depth boundaries with a generic network for
  disparity, optical flow or scene flow estimation.
\newblock In {\em {ECCV} 2018}.

\bibitem[\protect\citeauthoryear{Kar, H{\"{a}}ne, and
  Malik}{2017}]{DBLP:conf/nips/KarHM17}
Kar, A.; H{\"{a}}ne, C.; and Malik, J.
\newblock 2017.
\newblock Learning a multi-view stereo machine.
\newblock In {\em {NIPS} 2017}.

\bibitem[\protect\citeauthoryear{Kendall \bgroup et al\mbox.\egroup
  }{2017}]{DBLP:conf/iccv/KendallMDH17}
Kendall, A.; Martirosyan, H.; Dasgupta, S.; and Henry, P.
\newblock 2017.
\newblock End-to-end learning of geometry and context for deep stereo
  regression.
\newblock In {\em {ICCV} 2017}.

\bibitem[\protect\citeauthoryear{Kingma and
  Ba}{2015}]{DBLP:journals/arxiv/KingmaB14}
Kingma, D.~P., and Ba, J.
\newblock 2015.
\newblock Adam: {A} method for stochastic optimization.
\newblock In {\em {ICLR} 2015}.

\bibitem[\protect\citeauthoryear{Li \bgroup et al\mbox.\egroup
  }{2017}]{DBLP:journals/tvcg/LiSWZ17}
Li, D.; Shao, T.; Wu, H.; and Zhou, K.
\newblock 2017.
\newblock Shape completion from a single {RGBD} image.
\newblock {\em {IEEE} Trans. Vis. Comput. Graph.} 23(7):1809--1822.

\bibitem[\protect\citeauthoryear{Mayer \bgroup et al\mbox.\egroup
  }{2016}]{DBLP:conf/cvpr/MayerIHFCDB16}
Mayer, N.; Ilg, E.; H{\"{a}}usser, P.; Fischer, P.; Cremers, D.; Dosovitskiy,
  A.; and Brox, T.
\newblock 2016.
\newblock A large dataset to train convolutional networks for disparity,
  optical flow, and scene flow estimation.
\newblock In {\em {CVPR} 2016}.

\bibitem[\protect\citeauthoryear{Nealen \bgroup et al\mbox.\egroup
  }{2006}]{DBLP:conf/graphite/NealenISA06}
Nealen, A.; Igarashi, T.; Sorkine, O.; and Alexa, M.
\newblock 2006.
\newblock Laplacian mesh optimization.
\newblock In {\em {SIGGRAPH} 2006}.

\bibitem[\protect\citeauthoryear{Newcombe, Lovegrove, and
  Davison}{2011}]{DBLP:conf/iccv/NewcombeLD11}
Newcombe, R.~A.; Lovegrove, S.; and Davison, A.~J.
\newblock 2011.
\newblock {DTAM:} dense tracking and mapping in real-time.
\newblock In {\em {ICCV} 2011}.

\bibitem[\protect\citeauthoryear{Richter and
  Roth}{2018}]{DBLP:conf/cvpr/Richter018}
Richter, S.~R., and Roth, S.
\newblock 2018.
\newblock Matryoshka networks: Predicting 3d geometry via nested shape layers.
\newblock In {\em {CVPR} 2018}.

\bibitem[\protect\citeauthoryear{Tatarchenko \bgroup et al\mbox.\egroup
  }{2019}]{DBLP:conf/cvpr/TatarchenkoRRLKB19}
Tatarchenko, M.; Richter, S.~R.; Ranftl, R.; Li, Z.; Koltun, V.; and Brox, T.
\newblock 2019.
\newblock What do single-view 3{D} reconstruction networks learn?
\newblock In {\em {CVPR} 2019}.

\bibitem[\protect\citeauthoryear{Tatarchenko, Dosovitskiy, and
  Brox}{2017}]{DBLP:conf/iccv/TatarchenkoDB17}
Tatarchenko, M.; Dosovitskiy, A.; and Brox, T.
\newblock 2017.
\newblock Octree generating networks: Efficient convolutional architectures for
  high-resolution 3{D} outputs.
\newblock In {\em {ICCV} 2017}.

\bibitem[\protect\citeauthoryear{Wu \bgroup et al\mbox.\egroup
  }{2015}]{DBLP:conf/cvpr/WuSKYZTX15}
Wu, Z.; Song, S.; Khosla, A.; Yu, F.; Zhang, L.; Tang, X.; and Xiao, J.
\newblock 2015.
\newblock {3D ShapeNets}: A deep representation for volumetric shapes.
\newblock In {\em {CVPR} 2015}.

\bibitem[\protect\citeauthoryear{Wu \bgroup et al\mbox.\egroup
  }{2017}]{DBLP:conf/nips/0001WXSFT17}
Wu, J.; Wang, Y.; Xue, T.; Sun, X.; Freeman, B.; and Tenenbaum, J.
\newblock 2017.
\newblock {MarrNet}: 3{D} shape reconstruction via 2.5{D} sketches.
\newblock In {\em {NIPS} 2017}.

\bibitem[\protect\citeauthoryear{Xiao \bgroup et al\mbox.\egroup
  }{2010}]{DBLP:conf/cvpr/XiaoHEOT10}
Xiao, J.; Hays, J.; Ehinger, K.~A.; Oliva, A.; and Torralba, A.
\newblock 2010.
\newblock {SUN} database: Large-scale scene recognition from abbey to zoo.
\newblock In {\em {CVPR} 2010}.

\bibitem[\protect\citeauthoryear{Xie \bgroup et al\mbox.\egroup
  }{2019}]{DBLP:conf/iccv/XieHXSS19}
Xie, H.; Yao, H.; Sun, X.; Zhou, S.; and Zhang, S.
\newblock 2019.
\newblock {Pix2Vox}: Context-aware 3{D} reconstruction from single and
  multi-view images.
\newblock In {\em {ICCV} 2019}.

\bibitem[\protect\citeauthoryear{Yang \bgroup et al\mbox.\egroup
  }{2018}]{DBLP:journals/pami/YangRMTW18}
Yang, B.; Rosa, S.; Markham, A.; Trigoni, N.; and Wen, H.
\newblock 2018.
\newblock Dense 3{D} object reconstruction from a single depth view.
\newblock {\em {TPAMI}} {DOI:} 10.1109/TPAMI.2018.2868195.

\bibitem[\protect\citeauthoryear{Zhang \bgroup et al\mbox.\egroup
  }{2018}]{DBLP:conf/nips/ZhangZZTF018}
Zhang, X.; Zhang, Z.; Zhang, C.; Tenenbaum, J.; Freeman, B.; and Wu, J.
\newblock 2018.
\newblock Learning to reconstruct shapes from unseen classes.
\newblock In {\em {NeurIPS} 2018}.

\bibitem[\protect\citeauthoryear{Zhang \bgroup et al\mbox.\egroup
  }{2019}]{DBLP:conf/cvpr/ZhangVRP19}
Zhang, F.; Prisacariu, V.~A.; Yang, R.; and Torr, P. H.~S.
\newblock 2019.
\newblock Ga-net: Guided aggregation net for end-to-end stereo matching.
\newblock In {\em {CVPR} 2019}.

\bibitem[\protect\citeauthoryear{Zhao, Gao, and
  Lin}{2007}]{DBLP:journals/vc/ZhaoGL07}
Zhao, W.; Gao, S.; and Lin, H.
\newblock 2007.
\newblock A robust hole-filling algorithm for triangular mesh.
\newblock {\em The Visual Computer} 23(12):987--997.

\end{thebibliography}

\end{document}